\documentclass[10pt,twocolumn,letterpaper]{article}

\usepackage{cvpr}
\usepackage{times}
\usepackage{epsfig}
\usepackage{graphicx}
\usepackage{amsmath}
\usepackage{amssymb}
\usepackage{cite}
\usepackage{multirow}
\usepackage{rotating}

\usepackage{pifont}
\usepackage{array,arydshln}
\usepackage{hhline}
\usepackage{mathtools}

\usepackage[pagebackref=true,breaklinks=true,letterpaper=true,colorlinks,bookmarks=false]{hyperref}

\iffalse
\newcommand{\abel}[1]{\textcolor{green}{\bf [AG: #1]}}
\newcommand{\dvd}[1]{\textcolor{blue}{\bf [DVD: #1]}}
\newcommand{\vitto}[1]{\textcolor{magenta}{\bf [VF: #1]}}
\newcommand{\todo}[1]{\textcolor{red}{\bf [TODO: #1]}}
\else
\newcommand{\abel}[1]{}
\newcommand{\dvd}[1]{}
\newcommand{\vitto}[1]{}
\newcommand{\todo}[1]{}
\fi

\newcommand{\cmark}{\ding{51}}
\newcommand{\vspF}{\vspace{-4mm}}
\newcommand{\vspT}{\vspace{-2mm}}
\newcommand{\vspO}{\vspace{-1mm}}

\DeclareMathOperator*{\argmax}{argmax}
\DeclareMathOperator*{\IoU}{IoU}

\cvprfinalcopy 

\def\cvprPaperID{952} 

\begin{document}


\title{\vspF \vspT Objects as context for detecting their semantic parts \vspF}

{\small
\author{Abel Gonzalez-Garcia\\
{\tt\small a.gonzalez-garcia@sms.ed.ac.uk}
\and
Davide Modolo\\
{\tt\small davide.modolo@gmail.com}
\and
Vittorio Ferrari\\
{\tt\small vferrari@staffmail.ed.ac.uk}
\and
University of Edinburgh\\
}
}

\maketitle
\pagestyle{empty}

\begin{abstract}
  \vspF
  We present a semantic part detection approach that effectively leverages object information.
  We use the object appearance and its class
  as indicators of what parts to expect.
  We also model the expected relative location of parts inside the objects based on their appearance.
  We achieve this with a new network module, called OffsetNet, that efficiently predicts a variable number of part locations within a given object.
  Our model incorporates all these cues to detect parts in the context of their objects.
  This leads to considerably higher performance for the challenging task of part detection compared to using part appearance alone (+5 mAP on the PASCAL-Part dataset).
  We also compare to other part detection methods on both PASCAL-Part and CUB200-2011 datasets. 
\end{abstract}

\vspT
\section{Introduction}
\vspO

Semantic parts play an important role in visual recognition. They offer many advantages such as lower intra-class variability than whole objects, higher robustness to pose variation, and their configuration provides useful information about the aspect of the object. 
For these reasons, part-based models have gained attention for tasks such as fine-grained recognition~\cite{simon15iccv,liu12eccv,parkhi12cvpr,huang16cvpr,xiao15cvpr,lin15cvpr,zhang14eccv,zhang16cvpr,gavves13iccv,goering14cvpr}, object class detection and segmentation~\cite{chen14cvpr,azizpour12eccv,wang15iccv}, or attribute prediction~\cite{gkioxari15iccv,zhang13iccv,vedaldi14cvpr}.
Moreover, part localizations deliver a more comprehensive image understanding, enabling reasoning about object-part interactions in semantic terms.
Despite their importance, many part-based models detect semantic parts based only on their local appearance~\cite{parkhi12cvpr,chen14cvpr,liu12eccv,zhang13iccv,gkioxari15iccv,simon15iccv,xiao15cvpr,huang16cvpr,yang15iccv}.
While some works~\cite{lin15cvpr,zhang14eccv,zhang16cvpr,azizpour12eccv,vedaldi14cvpr,gavves13iccv,goering14cvpr}
leverage other types of information, they use parts mostly as support for other tasks. Part detection is rarely their focus.
Here we take part detection one step further and provide a specialized approach that exploits the unique nature of this task.

\begin{figure}
  \begin{center}
    \includegraphics[width=0.9\columnwidth]{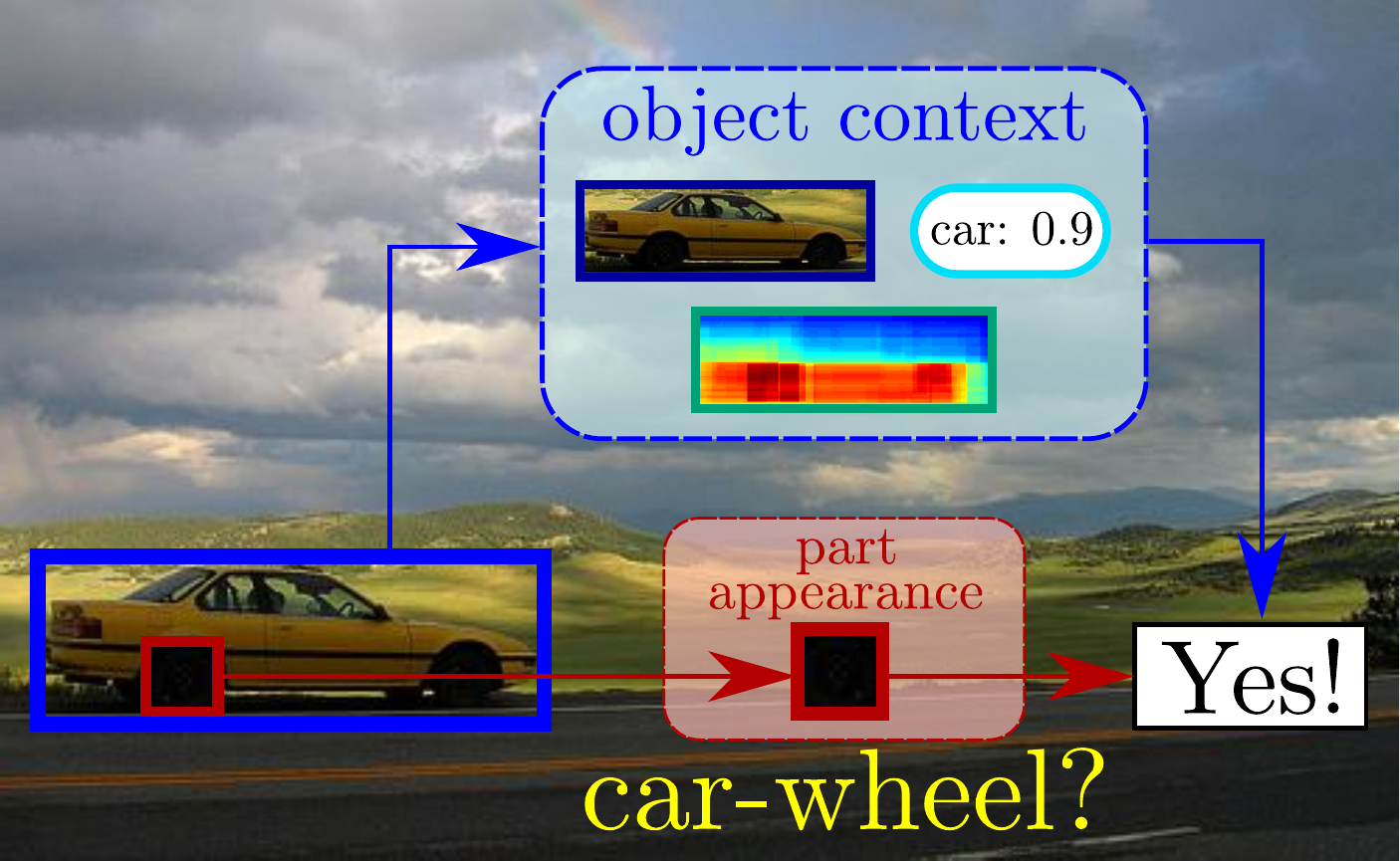}
\end{center}
\vspT
\vspO
  \caption{\small \textbf{Motivation for our model.} \it Part appearance alone might not be sufficiently discriminative in some cases. Our model uses object context to resolve ambiguities and help part detection. \vspF \vspO }
\label{fig:motivation}
\end{figure}

Parts are highly dependent on the objects that contain them. Hence, objects provide valuable cues to help detecting parts, creating an advantage over detecting them independently.
First, the class of the object gives a firm indication of what parts should be inside it, i.e.~only those belonging to that object class. 
For example, a dark round patch should be more confidently classified as a wheel if it is on a car, rather than on a dog (fig.~\ref{fig:motivation}). 
Furthermore, by looking at the object appearance we can determine in greater detail which parts might be present.
For example, a profile view of a car suggests the presence of a car door, and the absence of the licence plate. 
This information comes mostly through the viewpoint of the object, but also from other factors, such as the type of object (e.g. van), or whether the object is truncated (e.g. no wheels if the lower half is missing).
Second, objects also provide information about the location and shape of the parts they contain.
Semantic parts appear in very distinctive locations within objects, especially given the object appearance.
Moreover, they appear in characteristic relative sizes and aspect ratios. 
For example, wheels tend to be near the lower corners of car profile views, often in a square aspect ratio, and appear rather small.

In this work, we propose a dedicated part detection model that leverages all of the above object information.
We start from a popular Convolutional Neural Network (CNN) detection model~\cite{girshick15iccv}, which considers the appearance of local image regions only.
We extend this model to incorporate object information that complements part appearance by providing context in terms of object appearance, class and the relative locations of parts within the object.

We evaluate our part detection model on all 16 object classes in the PASCAL-Part dataset~\cite{chen14cvpr}.
We demonstrate that adding object information is greatly beneficial for the difficult task of part detection, leading to considerable performance improvements. 
Compared to a baseline detection model that considers only the local appearance of parts, our model achieves a +5 mAP improvement.
We also compare to methods that report part localization in terms of bounding-boxes~\cite{chen14cvpr,gkioxari15iccv,zhang14eccv,lin15cvpr,zhang16cvpr} on PASCAL-Part and CUB200-2011~\cite{WahCUB_200_2011}.
We outperform~\cite{chen14cvpr,gkioxari15iccv,zhang14eccv,lin15cvpr} and match the performance of~\cite{zhang16cvpr}.
We achieve this by an effective combination of the different object cues considered, demonstrating their complementarity.
Moreover our approach is general as it works for a wide range of object classes: we demonstrate it on 16 classes, as opposed to 1-7 in~\cite{chen14cvpr,parkhi12cvpr, wang15iccv,gkioxari15iccv,zhang14eccv,zhang14cvpr,lin15cvpr,zhang16cvpr, liu14eccv,hariharan15cvpr,sun11iccv_art, liang16eccv,ukita12cvpr,yang16cvpr,xia16eccv,gavves13iccv,goering14cvpr,zhang13iccv,simon15iccv,xiao15cvpr, xia17cvpr} (only animals and person).
Finally, we perform fully automatic object and part detection, without using ground-truth object locations at test time~\cite{chen14cvpr,wang15iccv,lin15cvpr,zhang16cvpr}. 
We released code for our method at~\cite{gonzalez-garcia18cvpr-code}.

\vspO
\section{Related work}
\vspO

\paragraph{DPM-based part-based models.}
The Deformable Part Model (DPM)~\cite{felzenszwalb10pami} detects objects as collections of parts, which are localized by local part appearance using HOG~\cite{dalal05cvpr} templates.
Most models based on DPM~\cite{felzenszwalb10pami,pandey11iccv,endres13cvpr,Ott11cvpr,nguyen15acomp,sadeghi2014eccv,divvala2012eccv,drayer14eccv,pedersoli11cvpr} consider parts as any image patch that is discriminative for the object class.
In our work instead we are interested in {\em semantic parts}, i.e. an object region interpretable and nameable by humans (e.g.`saddle'). 

Among DPM-based works,~\cite{azizpour12eccv} is especially related as they also simultaneously detect objects and their semantic parts. Architecturally, our work is very different:~\cite{azizpour12eccv} builds on DPM~\cite{felzenszwalb10pami}, whereas our model is based on modern CNNs and offers a tighter integration of part appearance and object context.
Moreover, the focus of~\cite{azizpour12eccv} is object detection, with part detection being only a by-product.
They train their model to maximize object detection performance, and thus they require parts to be located only roughly near their ground-truth box.
This results in inaccurate part localization at test time, as confirmed by the low part localization results reported~(table 5 of~\cite{azizpour12eccv}).
Finally, they only localize those semantic parts that are discriminative for the object class.
Our model, instead, is trained for precise part localization and detects all object parts.

\vspF
\paragraph{CNN-based part-based models.}
In recent years, CNN-based representations 
are quickly replacing hand-crafted features~\cite{dalal05cvpr,lowe04ijcv} in many domains, including semantic part-based models~\cite{bulat16eccv,chen14nips,gkioxari15iccv,hariharan15cvpr,huang16cvpr,liang16eccv,wang15cvpr,wang15iccv,xia16eccv,zhang14eccv,yang15iccv,simon15iccv,xiao15cvpr,modolo17pami}.
Our work is related to those that explicitly train CNN models to localize semantic parts using bounding-boxes~\cite{gkioxari15iccv,zhang14eccv,lin15cvpr,zhang16cvpr}, as opposed to keypoints~\cite{huang16cvpr,simon15iccv} or segmentation masks~\cite{hariharan15cvpr,liang16eccv,wang15cvpr,wang15iccv,xia16eccv,yang15iccv}. 
Many of these works~\cite{gkioxari15iccv,xiao15cvpr,simon15iccv,huang16cvpr,yang15iccv}  detect the parts used in their models based only on local part appearance, independently of their objects.
Moreover, they use parts as a means for object or action and attribute recognition, they are not interested in part detection itself.

Several fine-grained recognition works~\cite{gavves13iccv,goering14cvpr,zhang16cvpr} use nearest-neighbors to transfer part location annotations from training objects to test objects. 
They do not perform object detection, as ground-truth object bounding-boxes are used at both training and test time.  Here, instead, at test time we jointly detect objects and their semantic parts.

A few works~\cite{zhang14eccv,wang15iccv,xia17cvpr} use object information to refine part detections as a post-processing step.
Part-based R-CNN~\cite{zhang14eccv} refines R-CNN~\cite{girshick14cvpr} part detections by using nearest-neighbors from training samples.
Our model integrates object information also within the network, which allows us to deal with several object classes simultaneously, as opposed to only one~\cite{zhang14eccv}.
Additionally, we refine part detections with a new network module, OffsetNet, which is more efficient than nearest-neighbors.
Xia et al.~\cite{xia17cvpr} refine person part segmentations using the estimated pose of the person.
We propose a more general method that gathers information from multiple complementary object cues.
The method of~\cite{wang15iccv} is demonstrated only on 5 very similar classes from PASCAL-Part~\cite{chen14cvpr} (all quadrupeds), and on fully visible object instances from a manually selected subset of the test set (10\% of the full test set).
Instead, we show results on 105 parts over all 16 classes of PASCAL-Part, using the entire dataset.
Moreover,~\cite{wang15iccv} uses manually defined object locations at test time, whereas we detect both objects and their parts fully automatically at test time.


\section{Method}
\vspO

\label{sec:method}
\begin{figure*}
  \vspF
  \begin{center}
    \includegraphics[width=\textwidth]{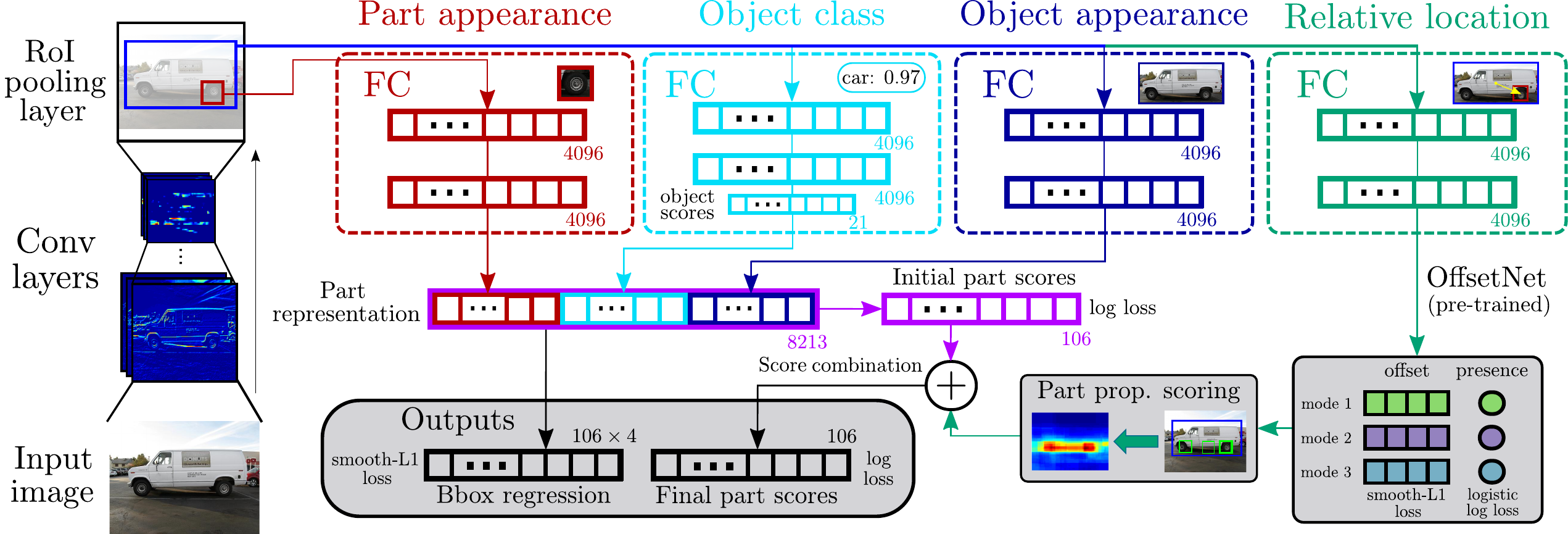}
\end{center}
\vspT
\caption{\small \textbf{Overview of our part detection model.} \it The model operates on part and object region proposals, passing them through several branches, and outputs part classification scores and regressed bounding-boxes.
  This example depicts the relative location branch for only object class \emph{car}. In practice, however, it processes all object classes simultaneously. When not explicitly shown, a small number next to the layer indicates its dimension for the PASCAL-Part~\cite{chen14cvpr} case, with a total of 20 object classes and 105 parts. \vspT } 
\label{fig:overview}
\end{figure*}

We define a new detection model specialized for parts which takes into account the context provided by the objects that contain them. 
This is the key advantage of our model over traditional part detection approaches, which detect parts based on their local appearance alone, independently of the objects~~\cite{gkioxari15iccv,simon15iccv,huang16cvpr,yang15iccv}.
We build on top of a baseline part detection model (sec.~\ref{sec:baseline}) and include various cues based on object class (sec,~\ref{sec:objSem}), object appearance (sec.~\ref{sec:objApp}), and the relative location of parts on the object (sec.~\ref{sec:relLoc}).
Finally, we combine all these cues to achieve more accurate part detections (sec.~\ref{sec:scoreComb}).

\vspF
\paragraph{Model overview.}
Fig.~\ref{fig:overview} gives an overview of our model. 
First, we process the input image through a series of convolutional layers.
Then, the Region of Interest (RoI) pooling layer produces feature representations from two different kind of region proposals, one for parts (red) and one for objects (blue).
Each part region gets associated with a particular object region that contains it (sec.~\ref{sec:supporting-proposal}).
Features for part regions are passed on to the part appearance branch, which contains two Fully Connected (FC) layers (sec.~\ref{sec:baseline}).
Features for object regions are sent to both the object class (sec.~\ref{sec:objSem}) and object appearance (sec.~\ref{sec:objApp}) branches, with three and two FC layers, respectively. 

For each part proposal, we concatenate the output of the part appearance branch with the outputs of the two object branches for its associated object proposal. We pass this refined part representation (purple) on to a part classification layer and a bounding-box regression layer (sec.~\ref{sec:scoreComb}).

Simultaneously, the relative location branch (green) also produces classification scores for each part region based on its relative location within the object (sec.~\ref{sec:relLoc}).
We combine the above part classification scores with those produced by relative location (big $+$ symbol, sec.~\ref{sec:scoreComb}), obtaining the final part classification scores. The model outputs these and regressed bounding-boxes.

\vspT
\subsection{Baseline model: part appearance only}
\vspO
\label{sec:baseline}

As baseline model we use the popular Fast R-CNN~\cite{girshick15iccv}, which was originally designed for object detection.
It is based on a CNN that scores a set of region proposals~\cite{uijlings13ijcv} by processing them through several layers of different types.
The first layers are convolutional and they process the whole image once.
Then, the RoI pooling layer extracts features for each region proposal, which are later processed by several FC layers.
The model ends with two sibling output layers, one for classifying each proposal into a part class, and one for bounding-box regression, which refines the proposal shape to match the extent of the part more precisely. 
The model is trained using a multi-task loss which combines these two objectives.
This baseline corresponds to the part appearance branch in fig.~\ref{fig:overview}.

We follow the usual approach~\cite{girshick15iccv} of fine-tuning for the used dataset on the current task, part detection, starting from a network pre-trained for image classification~\cite{krizhevsky12nips}. 
The classification layer of our baseline model has as many outputs as part classes, plus one output for a generic background class.
Note how we have a single network for all part classes in the dataset, spanning across all object classes.

\subsection{Adding object appearance and class}
\vspO
\label{sec:objAppSem}

The baseline model tries to recognize parts based only on the appearance of individual region proposals.
In our first extension, we include object appearance and class information by integrating it inside the network.
We can see this as selecting an adequate contextual spatial support for the classification of each proposal into a part class.

\vspF
\subsubsection{Supporting proposal selection}
\begin{figure}[t]
 \vspT 
 \vspO 
  \begin{center}
    \includegraphics[width=0.9\columnwidth]{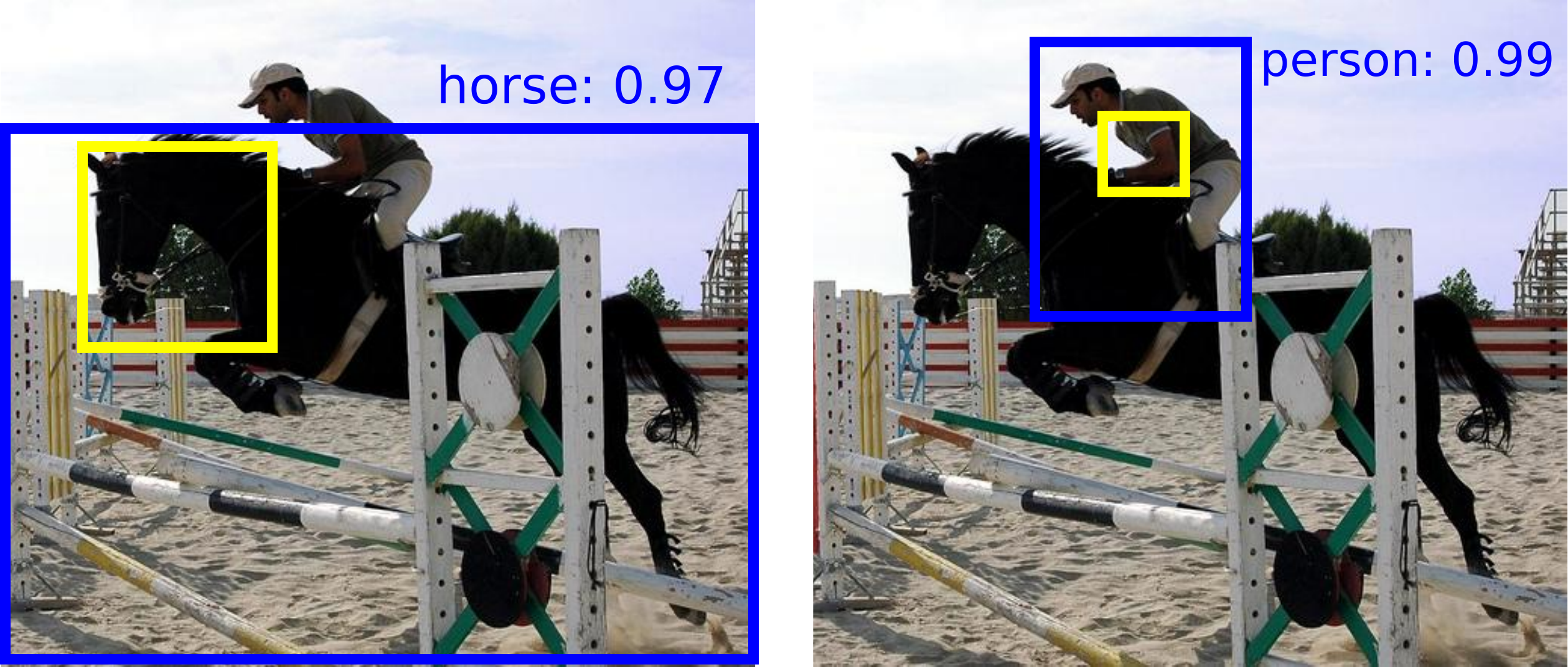}
\end{center}
\vspT
  \caption{\small \textbf{Examples of supporting proposal selection.} \it For each part proposal (yellow), we select as its supporting proposal (blue) the highest scored among the object proposals that contain it.\vspF } 
\label{fig:superproposals}
\end{figure}

\vspO
\label{sec:supporting-proposal}
Our models use two types of region proposals (sec.~\ref{sec:implementation}).
{\em Part proposals} are candidate regions that might cover parts.
Analogously, {\em object proposals} are candidates to cover objects.
The baseline model uses only part proposals.
In our models, instead, each part proposal $p$ is accompanied by a \emph{supporting object proposal} $S_{sup}(p)$, which must fulfill two requirements (fig.~\ref{fig:superproposals}).
First, it needs to contain the part proposal, i.e. at least 90\% of $p$ must be inside $S_{sup}(p)$.
Second, it should tightly cover the object that contains the part, if any.
For example, if the part proposal is on a wheel, the supporting proposal should be on the car that contains that wheel.
To achieve this, we select the highest scored proposal among all object proposals containing $p$, where the score is the object classification score for any object class.

Formally, let $p$ be a part proposal and $S(p)$ the set of object proposals that contain $p$.
Let $\phi_{obj}^k(S_n)$ be the classification score of proposal $S_n\in S(p)$ for object class $k$.
These scores are obtained by first passing all object proposals through three FC layers as in the object detector~\cite{girshick15iccv}. 
We select the supporting proposal $S_{sup}(p)$ for $p$ as
\begin{equation}
  \vspT
  S_{sup}(p) = \argmax_{S_n\in S(p)}\big[ \max_{k\in \{1,...,K\} }\phi_{obj}^k(S_n) \big],
  \label{eq:suppObj}
\end{equation}
where $K$ is the total number of object classes in the dataset.

\vspT
\subsubsection{Object class}
\vspO
\label{sec:objSem}
The class of the object provides cues about what part classes might be inside it. 
For example, a part proposal on a dark round patch cannot be confidently classified as a wheel based solely on its appearance (fig.~\ref{fig:motivation}). 
If the corresponding supporting object proposal is a car, the evidence towards it being a wheel grows considerably.
On the other hand, if the supporting proposal is a dog, the patch should be confidently classified as {\em not} a wheel.

Concretely, we process convolutional features pooled from the supporting object proposal through three FC layers (fig.~\ref{fig:overview}). The third layer performs object classification and outputs scores for each object class, including a generic background class. These scores can be seen as object semantic features, which complement part appearance.

\vspF
\subsubsection{Object appearance} 
\vspT
\label{sec:objApp}

The appearance of the object might bring even more detailed information about what part classes it might contain.
For example, the side view of a car indicates that we can expect to find wheels, but not a licence plate.
We model object appearance by processing the convolutional features of the supporting proposal through two FC connected layers (fig.~\ref{fig:overview}). 
This type of features have been shown to successfully capture the appearance of objects~\cite{donahue14icml,ge15cvpr}.

\vspO
\subsection{Adding relative location}
\vspO
\label{sec:relLoc}

\begin{figure}[t]
  \begin{center}
    \vspF
    \vspO
    \includegraphics[width=\columnwidth]{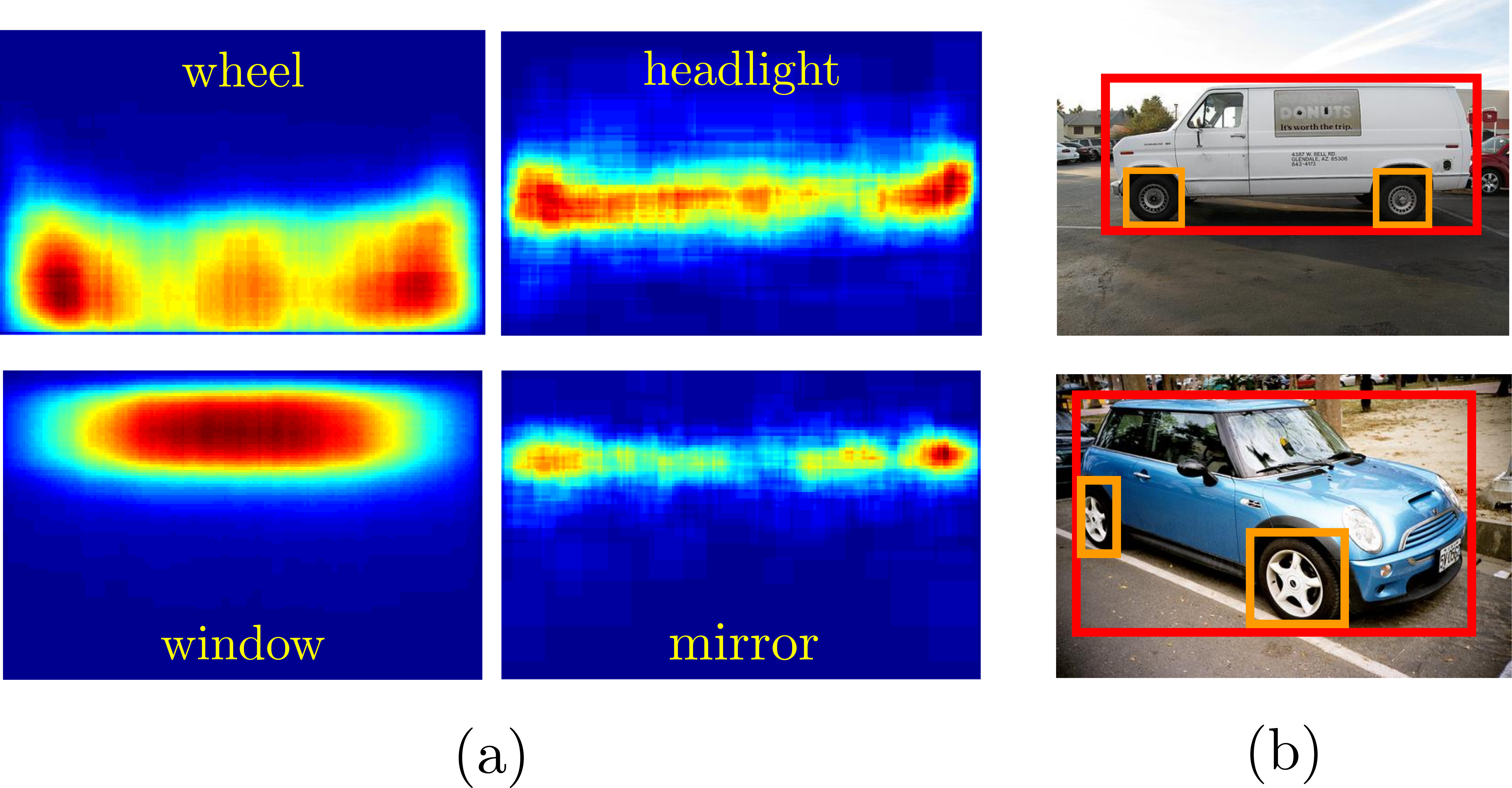}
\end{center}
\vspF
\caption{\small \textbf{Prior distributions and examples of part relative locations.} \it (a) Heatmaps created using part ground-truth bounding-boxes normalized to the object bounding-box, using all \emph{car} training samples. (b) Examples of part ground-truth bounding-boxes inside object bounding-boxes of class \emph{car}, in different viewpoints. 
  \vspF }
\label{fig:relLocPrior}
\end{figure}

We now add another type of information that could be highly beneficial: the relative location of the part with respect to the object.
Parts appear in very distinct and characteristic relative locations and sizes within the objects.
Fig.~\ref{fig:relLocPrior}{\color{red}a} shows examples of prior relative location distributions for some part classes as heatmaps. 
These are produced by accumulating all part ground-truth bounding-boxes from the training set, in the normalized coordinate frame of the bounding-box of their object.
Moreover, this part location distribution can be sharper if we condition it on the object appearance, especially its viewpoint.
For example, the \emph{car-wheel} distribution on profile views of cars will only have two modes (fig.~\ref{fig:relLocPrior}{\color{red}b}) instead of the three shown in fig.~\ref{fig:relLocPrior}{\color{red}a}.

\vspF
\paragraph{Overview.}
Our relative location model is specific to each part class within each object class (e.g. a model for car-wheel, another model for cat-tail). Below we explain the model for one particular object and part class.
Given an object proposal $o$ of that object class, our model suggests windows for the likely position of each part inside the object.
Naturally, these windows will also depend on the appearance of $o$. 
For example, given a car profile view, our model suggests square windows on the lower corners as likely to contain wheels (fig.~\ref{fig:relLocPrior}{\color{red}b} top). 
Instead, an oblique view of a car will also suggest a wheel towards the lower central region, as well as a more elongated aspect ratio for the wheels on the side (fig.~\ref{fig:relLocPrior}{\color{red}b} bottom).
We generate the suggested windows using a special kind of CNN, which we dub \emph{OffsetNet} (see fig.~\ref{fig:overview}, Relative location branch).
Finally, we score each part proposal according to its overlap with the suggested windows. This indicates the probability that a proposal belongs to a certain part class, based on its relative location within the object, and on the object appearance (but it does not depend on part appearance).

\vspF
\vspO
\paragraph{OffsetNet Model.}

OffsetNet directly learns to regress from the appearance of an object proposal $o$ to the relative location of a part class within it.
Concretely, it learns to produce a 4D offset vector $\delta v$ that points to the part inside $o$. 
In fact, OffsetNet produces a set of vectors $\Delta v$, as some objects have multiple instances of the same part inside (e.g. cars with multiple wheels).
Intuitively, a CNN is a good framework to learn this regressor, as the activation maps of the network contain localized information about the parts of the object~\cite{simon14accv,zeiler14eccv,gonzalez-garcia17ijcv}.

OffsetNet generates each offset vector in $\Delta v$ through a regression layer.
To enable OffsetNet to output multiple vectors we build multiple parallel regression layers.
We set the number of parallel layers to the number of modes of the prior distribution for each part class (fig.~\ref{fig:relLocPrior}{\color{red}}).
For example, the prior \emph{car-wheel} has three modes, leading to three offset regression layers in OffsetNet (fig.~\ref{fig:overview}).
On the other hand, OffsetNet only has one regression layer for \emph{person-head}, as its prior distribution is unimodal.

In some cases, however, not all modes are active for a particular object instance (e.g. profile views of cars only have two active modes out of the three, fig.~\ref{fig:relLocPrior}{\color{red}b}).
For this reason, each regression layer in OffsetNet has a sibling layer that predicts the presence of that mode in the input detection $o$, and outputs a presence score $\rho$.
This way, even if the network outputs multiple offset vectors, only those with a high presence score will be taken into account. This construction effectively enables OffsetNet to produce a variable number of output offset vectors, depending on the input $o$.

\begin{figure}[t]
  \vspT
  \begin{center}
    \includegraphics[width=\columnwidth]{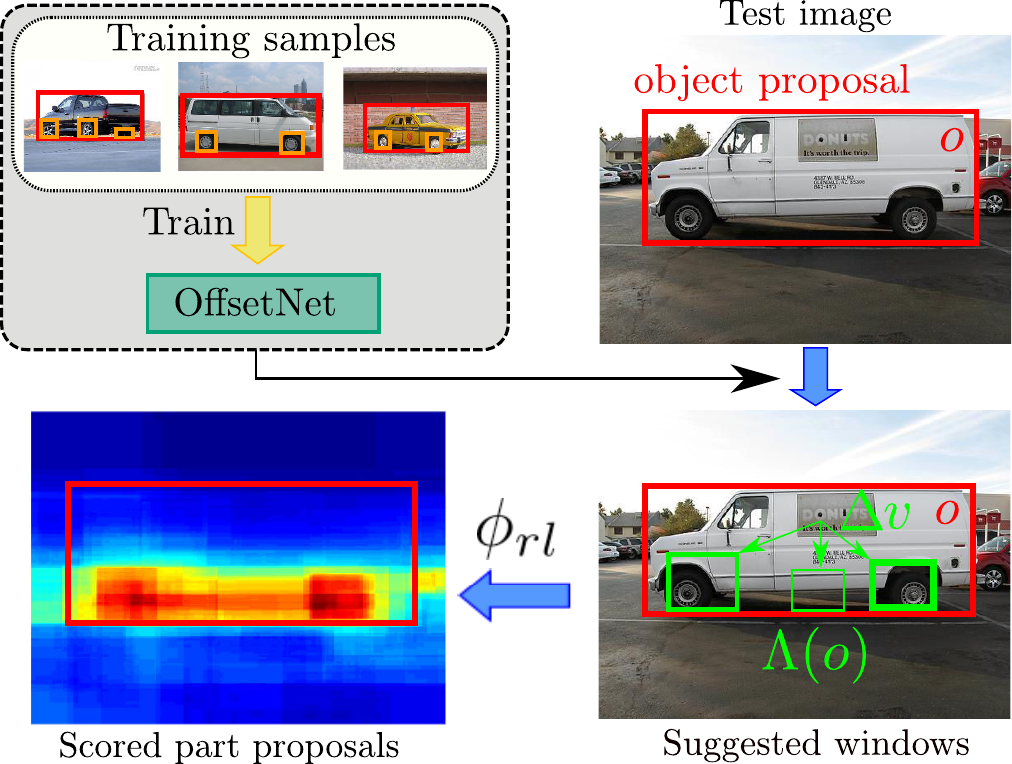}
\end{center}
\vspT
\caption{\small \textbf{Example of scoring part proposals based on relative location.} \it Part class: \emph{car-wheel}. Each car detection suggests windows likely to contain car-wheel within it. We score part proposals by computing their max IoU with any suggested window, and weighting them by their presence score and the object detection score.
We show only one \emph{car} detection for clarity.
\vspF \vspT }
\label{fig:nnApproach}
\end{figure}

\vspF
\vspO
\paragraph{Training OffsetNet.}
We train one OffsetNet for all part classes simultaneously by arranging $N$ parallel regression layers, where $N$ is the maximum number of modes over all part classes (4 for PASCAL-Part). 
If a part class has fewer than N modes, we simply ignore its regression output units in the extra layers.
We train the offset regression layers using a smooth-L1 loss, on the training samples described in sec.~\ref{sec:implementation} (analog to the bounding-box regression of Fast R-CNN~\cite{girshick15iccv}).
We train the presence score layer using a logistic log loss: $L(x,c) = \log(1+e^{-cx})$, where $x$ is the score produced by the network, and $c$ is a binary label indicating whether the current mode is present ($c=+1$) or not ($c=-1$). We generate $c$ using annotated ground-truth bounding-boxes (sec.~\ref{sec:implementation}).
This loss implicitly normalizes score $x$ using the sigmoid function. 
After training, we add a sigmoid layer to explicitly normalize the output presence score: $\rho = 1/(1+e^{-x}) \in [0,1]$.

\begin{figure}
  \vspO
  \begin{center}
    \includegraphics[width=\columnwidth]{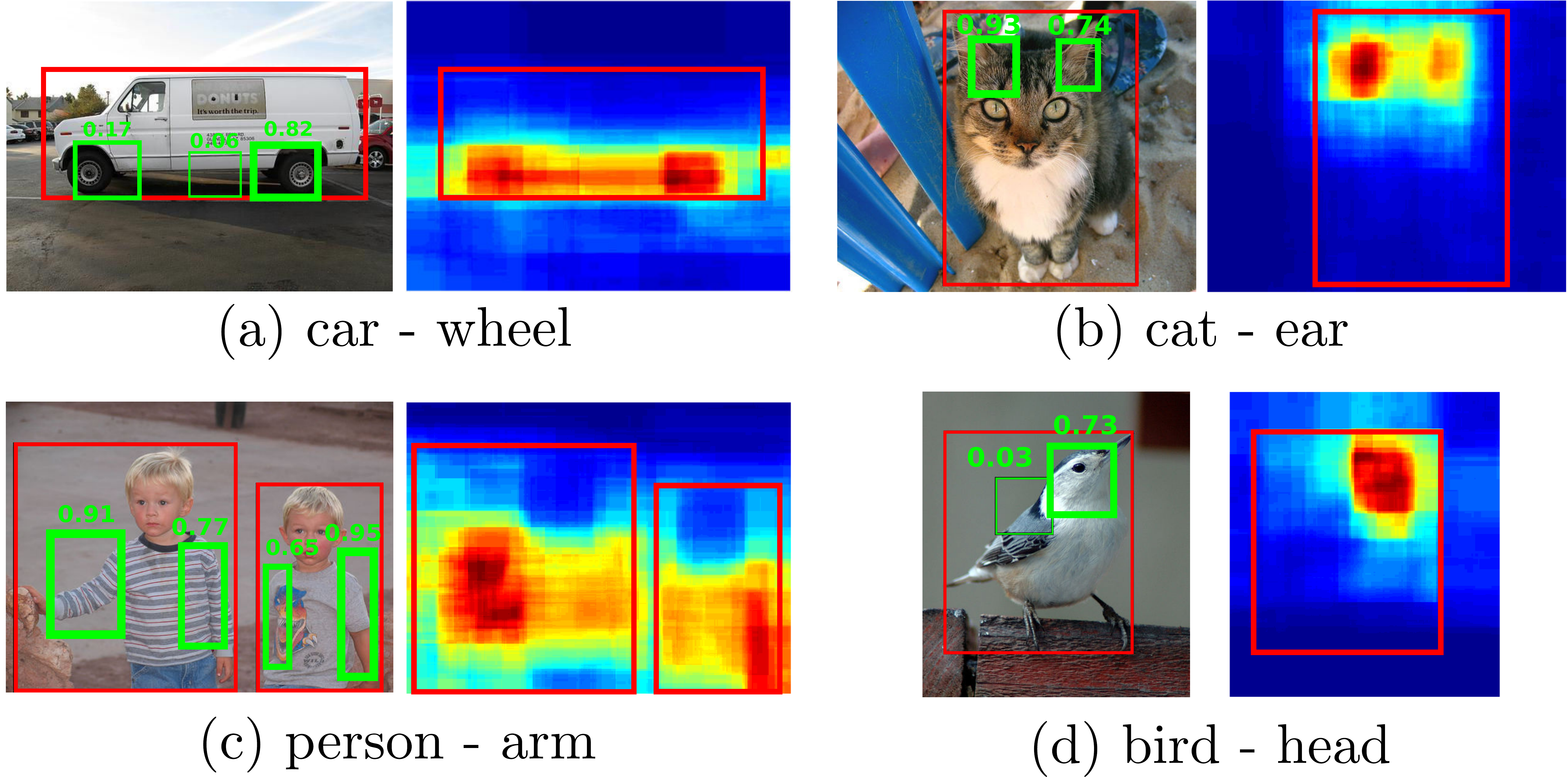}
\end{center}
\vspF
\caption{\small \textbf{OffsetNet results.} \it Examples of windows suggested by OffsetNet (green) for different part classes, given the input object detections (red). We also show the heatmap generated by scoring all part proposals, showing how highly scored proposals occupy areas likely to contain part instances. The presence scores clearly indicate which suggested windows should be relied on (e.g. the second head of the bird and the middle wheel of the van have very low scores and are discounted in $\phi_{rl}$).\vspF }
\label{fig:windowNetRes}
\end{figure}

\vspF
\vspO
\paragraph{Generating suggested windows.}
At test time, given an input object proposal $o$, OffsetNet generates $M$ pairs $\{(\delta v_i, \rho_i)\}_{i=1}^{M}$ of offset vectors $\delta v_i \in \Delta v$ and presence scores $\rho_i$ for each part class, where $M$ is the number of modes in the prior distribution.
We apply the offset vectors $\delta v_i$ to $o$, producing a set of suggested windows $\Lambda(o) = \{ o + \delta v_i \}_{i=1}^M = \{ w_i \}_{i=1}^M$.

\vspT
\paragraph{Scoring part proposals.}
At test time, we score all part proposals of an image by generating windows with OffsetNet for all the detected objects in the image.
Let $O$ be a set of object detections in the image, i.e. object proposals with high score after non-maxima suppression~\cite{felzenszwalb10pami}. We produce these automatically using standard Fast R-CNN~\cite{girshick15iccv}.
Let $\phi_{obj}(o)$ be the score of detection $o\in O$ for the considered object class.
We compute the relative location score $\phi_{rl}(p)$ for part proposal $p$ using its overlap with all windows suggested by OffsetNet 
\begin{equation}
  \vspT
  \label{eq:scoreRL}
  \phi_{rl}(p) = \max_{w_i\in\Lambda(o), o \in O}(\IoU(p,w_i)\cdot \rho_i\cdot \phi_{obj}(o)),
\end{equation}
where we use Intersection-over-Union (IoU) to measure overlap.
Here, $\Lambda(o) = \{ w_i \}_{i=1}^M$ is the set of suggested windows output by OffsetNet for object detection $o$, and $\rho_i$ is the associated presence score for each individual window $w_i$.
Suggested windows with higher presence score have higher weight in the overall relative location score $\phi_{rl}$.
The score of the object detection $\phi_{obj}(o)$ is also used to weight all suggested windows based on it.
Consequently, object detections with higher score provide stronger cues through higher relative location scores $\phi_{rl}$.
Fig.~\ref{fig:nnApproach} depicts this scoring procedure.

Fig.~\ref{fig:windowNetRes} shows examples of windows suggested by OffsetNet, along with their presence score and a heatmap generated by scoring part proposals using eq.~\eqref{eq:scoreRL}. We can see how the suggested windows cover very likely areas for part instances on the input objects, and how the presence scores are crucial to decide which windows should be relied on.

\vspO
\subsection{Cue combination}
\vspT
\label{sec:scoreComb}
We have presented multiple cues that can help part detection.
These cues are complementary, so our model needs to effectively combine them.

We concatenate the output of the part appearance, object class and object appearance branches
and pass them on to a part classification layer that combines them and produces initial part classification scores (purple in fig.~\ref{fig:overview}). 
Therefore, we effectively integrate object context into the network, resulting in the automatic learning of object-aware part representations.
We argue that this type of context integration has greater potential than just a post-processing step~\cite{wang15iccv,zhang14eccv}.

The relative location branch, however, is special as its features have a different nature and much lower dimensionality (4 vs 4096). 
To facilitate learning, instead of directly concatenating them, this branch operates independently of the others and computes its own part scores.
Therefore, we linearly combine the initial part classification scores with those delivered by the relative location branch (big $+$ in fig.~\ref{fig:overview}).
For some part classes, the relative location might not be very indicative due to high variance in the training samples (e.g. \emph{cat-nose}).
In some other cases, relative location can be a great cue (e.g. the position of \emph{cow-torso} is very stable across all its instances). 
For this reason, we learn a separate linear combination for each part class.
We do this by maximizing part detection performance on the training set, using grid search on the mixing weight in the $[0,1]$ range. 
We define the measure of performance in sec.~\ref{sec:results}.

\vspO
\section{Implementation details}
\vspO
\label{sec:implementation}
\paragraph{Proposals.}
Object proposals~\cite{alexe12pami,dollar14eccv,uijlings13ijcv} are designed to cover whole objects, and sometimes fail to find small parts.
To alleviate this issue, we changed the standard settings of Selective Search~\cite{uijlings13ijcv}, by decreasing the minimum box size to 10. 
This results in adequate proposals even for parts: reaching 71.4\% 
recall with $\sim$3000 proposals (IoU $>$0.5).
For objects, we keep the standard settings (minimum box size 20), resulting in $\sim$2000 proposals.

\vspF
\vspO
\paragraph{Training the part detection network.}
Our networks are pre-trained on ILSVRC12~\cite{ilsvrc12} for image classification and fine-tuned on PASCAL-Part~\cite{chen14cvpr} for part detection, or on PASCAL VOC 2010~\cite{everingham10ijcv} for object detection, using MatConvNet~\cite{vedaldi15mm}.
Fine-tuning for object detection follows the Fast R-CNN procedure~\cite{girshick15iccv}.
For part detection fine-tuning we changed the following settings. 
Positive samples for parts overlap any part ground-truth $>0.6$ IoU, whereas negative samples overlap $<0.3$.
We train for 12 epochs with learning rate 10$^{-3}$, and then for 4 epochs with 10$^{-4}$.

We jointly train part appearance, object appearance, and object class branches for a multi-task part detection loss.
We modify the RoI pooling layer to pool convolutional features from both the part proposal and the supporting object proposal. 
Backpropagation through this layer poses a problem, as~\eqref{eq:suppObj} is not differentiable. 
We address this by backpropagating the gradients only through the area of the convolutional map covered by the object proposal selected by the $\argmax$.
We obtain the object scores used in~\eqref{eq:suppObj} from the object class branch, which is previously initialized using the standard Fast R-CNN object detection loss, in order to provide reliable object scores when joint training starts.

\vspF
\vspO
\paragraph{Training OffsetNet.}
We need object samples and part samples to train OffsetNet.
Our object samples are all object ground-truth bounding-boxes and object proposals with IoU $\geq$ 0.7 in the training set.
Our part samples are only part ground-truth bounding-boxes.
We split the horizontal axis in $M$ regions, where $M$ is the number of modes in the part class prior relative location distribution. 
We assign each part ground-truth bounding-box in the object to the closest mode.
If a mode has more than one part bounding-box assigned, we pick one at random.
In case a mode has no instance assigned (e.g. occluded wheel) for a particular training sample, the loss function omits the contribution of that mode.
All layers except the top ones are initialized with a Fast R-CNN network trained for object detection.
Similarly to the other networks, we train it for 16 epochs, but with learning rates 10$^{-4}$ and 10$^{-5}$.

\section{Results and conclusions}
\vspO
\label{sec:results}

\subsection{Validation of our model}
\vspO
\label{sec:resultsVal}
\paragraph{Dataset.}
We present results on PASCAL-Part~\cite{chen14cvpr}, which has pixel-wise part annotations for the images of PASCAL VOC 2010~\cite{everingham10ijcv}. For our experiments we fit a bounding-box to each part segmentation mask.
We pre-process the set of part classes as follows.
We discard additional information on semantic part annotations, such as `front' or `left' (e.g. both ``car wheel front left'' and ``car wheel back right'' become \emph{car-wheel}).
We merge continuous subdivisions of the same semantic part (``horse lower leg'' and ``horse upper leg'' become \emph{horse-leg}).
Finally, we discard tiny parts, with average width and height over the training set $\leq$15 pixels (e.g. ``bird eye''), and rare parts that appear $<10$ times (e.g. ``bicycle headlight''). 
After this pre-processing, we obtain a total of 105 part classes for 16 object classes.
We train our methods on the \texttt{train} set and test them on the \texttt{val} set (the \texttt{test} set is not annotated in PASCAL-Part).
We note how we are the first work to present fully automatic part detection results on the whole PASCAL-Part dataset.

\vspF
\paragraph{Performance measure.}
Just before measuring performance we remove duplicate detections using non-maxima suppression~\cite{felzenszwalb10pami}.
We measure part detection performance using Average Precision (AP), following the PASCAL VOC protocol~\cite{everingham10ijcv}.
We consider a part detection to be correct if its IoU with a ground-truth part bounding-box is $>0.5$.

\begin{table}[t]
\centering
\small
\resizebox{\columnwidth}{!}{
\begin{tabular}{|c|c | c : c : c | c| c|}
\hline
& \textbf{Model} & \textbf{Obj. App} & \textbf{Obj. Cls} & \textbf{Rel Loc} & \textbf{mAP} \\
\hline
\multirow{10}{*}{\begin{turn}{90}AlexNet\end{turn}} & Baseline~\cite{girshick15iccv} (part appearance only) & & & & 22.1 \\ \cdashline{2-6}  
& Obj. appearance & \cmark & & & 25.1  \\ 
& Obj. class & & \cmark & & 23.0\\ 
& Obj. app + cls & \cmark & \cmark & & 25.7  \\ \cdashline{2-6} 
& OffsetNet ($M=1$) & & & \cmark& 24.3  \\ 
& OffsetNet & & & \cmark & 24.7  \\ \cdashline{2-6} 
& Obj. app + OffsetNet & \cmark & & \cmark& 26.8  \\ 
& Full (Obj. app + cls + OffsetNet) & \cmark & \cmark & \cmark& \bf{27.4}   \\ 
\cline{2-6} 
& Baseline~\cite{girshick15iccv} (bbox-reg) & & & & 24.5 \\ \cdashline{2-6}  
& Full (bbox-reg) & \cmark & \cmark & \cmark& \bf{29.5}   \\
\hline
\multirow{7}{*}{\begin{turn}{90}VGG16 \end{turn}} & Baseline~\cite{girshick15iccv} (bbox-reg) & & & & 35.8 \\ \cdashline{2-6}   
& Obj. appearance (bbox-reg) & \cmark & & & 38.2  \\ 
& Obj. class (bbox-reg)& & \cmark & & 35.4\\ 
& Obj. app + cls (bbox-reg)& \cmark & \cmark & & 38.7  \\ \cdashline{2-6}   
& OffsetNet (bbox-reg)& & & \cmark & 38.5  \\ \cdashline{2-6}  
& Obj. app + OffsetNet (bbox-reg)& \cmark & & \cmark& 39.4  \\ 
& Full (bbox-reg) & \cmark & \cmark & \cmark& \bf{40.1}   \\
\hline

\end{tabular}}
\vspace{-1mm}
  \caption{\small \textbf{Part detection results on PASCAL-Part.} \it The first 10 rows use AlexNet, the last 7 use VGG16. The baseline model uses only part appearance. All other models include it too. \vspF } 
 \label{tab:resultsAlexNet}
\end{table}

\vspF
\paragraph{Baseline results.}
Tab.~\ref{tab:resultsAlexNet} presents part detection results. 
As base network in the top 10 rows we use AlexNet~\cite{krizhevsky12nips} (convolutional layers on the leftmost column of fig.~\ref{fig:overview}).
As a baseline we train a Fast R-CNN model to directly detect parts, using only their own appearance (sec.~\ref{sec:baseline}).
This achieves only 22.1 mAP (without bounding-box regression).
As a reference, the same model achieves 48.5 mAP, when trained and evaluated for object class detection on PASCAL VOC 2010~\cite{everingham10ijcv} (same object classes as PASCAL-Part).
This massive difference in performance demonstrates the inherent difficulty of the part detection task.

\vspF
\vspO
\paragraph{Adding object appearance and class.}
By adding object appearance (sec.~\ref{sec:objApp}), performance increases by +3 mAP, which is a significant improvement.
Adding object class (sec.~\ref{sec:objSem}) also helps, albeit less so (+0.9 mAP).
This indicates that the appearance of the object contains extra knowledge relevant for part discrimination (e.g. viewpoint), which the object class alone cannot provide.
Furthermore, the combination of both types gives a small additional boost (+0.6 mAP compared to using only object appearance).
Although in principle object appearance subsumes its class, having a more explicit and concise characterization of the class is beneficial for part discrimination.

\begin{figure*}
\vspT
  \begin{center}
    \includegraphics[width=\textwidth]{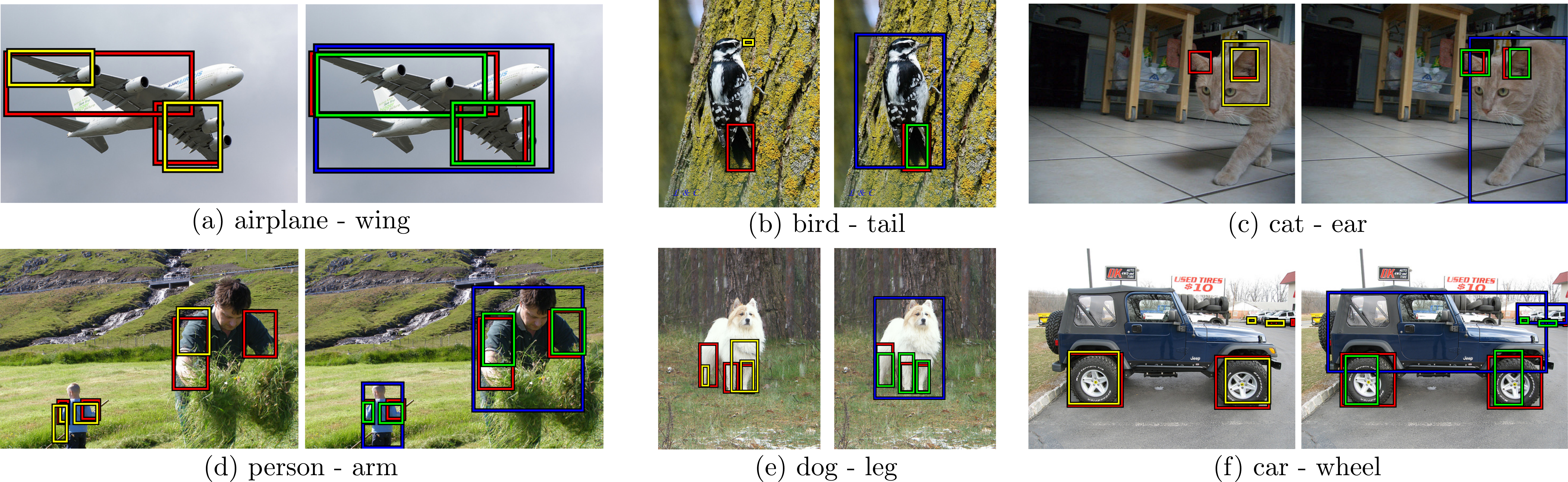}
\end{center}
\vspT
\caption{\small \textbf{Qualitative results.} \it Example part detections for the baseline model (yellow) and our model combining all cues (green). We also show part ground-truth bounding-boxes (red), and object detections output by our method (blue).\vspF } 
\label{fig:qualRes}
\end{figure*}

\vspF
\paragraph{Adding relative location.}
Our relative location model (OffsetNet, sec.~\ref{sec:relLoc}) also brings improvements.
We present results for two OffsetNet versions.
In the first one, we fix the number of modes $M$ to $1$ for all part classes, regardless of the complexity of the prior distribution.
In this simplified setting, OffsetNet already benefits part detection (+2.2 mAP over the baseline).
When setting $M$ based on the prior distribution as explained in sec.~\ref{sec:relLoc}, the improvement further rises to +2.6 mAP.

\vspF
\paragraph{Combining cues.}
Finally, we combine all our cues as in sec.~\ref{sec:scoreComb} (always using also part appearance).
First, we combine our relative location model with object appearance.
This combination is beneficial and surpasses each cue alone.

Our best model (full) combines all cues and achieves +5.3 mAP over the baseline. This as a substantial improvement when considering the high difficulty of the task, as demonstrated by the rather low baseline performance. 
This result shows that all cues we propose are complementary to part appearance and to each other; when combined, all contribute to the final performance.

We also tested the baseline and our model using bounding-box regression (bbox-reg in tab.~\ref{tab:resultsAlexNet}). 
This helps in all cases.
Importantly, the improvement brought by our full model over the baseline (+5 mAP) is similar to the one without bounding-box regression (+5.3 mAP).

\vspF
\vspO
\paragraph{Partial object information.}
We explore here whether our method still helps when the object is only partially visible.
We evaluate the baseline and our full method only on occluded objects, as indicated in the annotations of PASCAL-Part. 
The baseline gives 13.7 mAP, whereas our full method achieves 17.2 mAP.
This demonstrates that integrating object cues benefits part detection even on partially occluded objects.
One reason for this desirable behaviour is that our model learns to deal with occluded objects during training, as the training set of PASCAL-Part also includes such cases.

\vspF
\vspO
\paragraph{Example detections.}
Fig.~\ref{fig:qualRes} shows some part detection examples for both the baseline and our full model (without bounding-box regression).
In general, our model localizes parts more accurately, fitting the part extent more tightly (fig.~\ref{fig:qualRes}{\color{red}a},\ref{fig:qualRes}{\color{red}e}).
Moreover, it also finds some part instances missed by the baseline (fig.~\ref{fig:qualRes}{\color{red}b},~\ref{fig:qualRes}{\color{red}c}).
Our method uses object detections automatically produced by Fast R-CNN~\cite{girshick15iccv}. When these are inaccurate, our model can sometimes produce worse part detections than the baseline (fig.~\ref{fig:qualRes}{\color{red}f}).

\vspF
\vspO
\paragraph{Runtime.}
We report runtimes on a Titan X GPU.
The baseline takes 4.3s/im, our model 7.1s/im.
Note how we also output object detections, which the baseline does not.

\vspF
\vspO
\paragraph{Results for VGG16.}
We also present results for the deeper VGG16 network~\cite{simonyan15iclr} (last 7 rows in tab.~\ref{tab:resultsAlexNet}).
The relative performance of our model and the baseline is analog to the AlexNet case, but with higher mAP values.
The baseline achieves 35.8 mAP with bounding-box regression.
Our full model achieves 40.1 mAP, i.e. an improvement of magnitude comparable to the AlexNet case (+4.3 mAP).

\begin{table}[t]
\centering
\resizebox{\columnwidth}{!}{
\begin{tabular}{|c || c | c || c | c || c | c |}
  \hline
  \textbf{Comparison to } & \multicolumn{2}{c||}{\textbf{Chen et al.~\cite{chen14cvpr}}} & \multicolumn{2}{c||}{\textbf{Fine-grained~\cite{zhang14eccv}\cite{lin15cvpr}\cite{zhang16cvpr}} } &\multicolumn{2}{c|}{\textbf{Gkioxari et al.~\cite{gkioxari15iccv} }}\\
  \hline
  Dataset & \multicolumn{2}{c||}{PASCAL-Part} & \multicolumn{2}{c||}{CUB200-2011} & \multicolumn{2}{c|}{PASCAL VOC09}\\ 
  \hline
  Measure & POP & PCP & \multicolumn{2}{c||}{PCP} & AP (0.3) & AP (0.5)\\ 
  \hline
  Obj GT at test & \cmark & \cmark &  & \cmark & & \\ 
  \hline
  \multirow{3}{*}{Theirs} & \multirow{3}{*}{44.5} & \multirow{3}{*}{70.5} & \multirow{3}{*}{66.1~\cite{zhang14eccv}} & 74.0~\cite{zhang14eccv} & \multirow{3}{*}{38.7}  & \multirow{3}{*}{17.1} \\
  & & & & 85.0~\cite{lin15cvpr} & & \\  
  & & & & \textbf{94.2}~\cite{zhang16cvpr} & & \\  
  
  \hline
  Ours & \textbf{51.3} & \textbf{72.6} & \textbf{91.9} & 92.7 & \textbf{53.6 (65.5)} & \textbf{21.6 (44.7)} \\
 \hline
\end{tabular}}\vspace{1mm} 
  \caption{\small \textbf{Comparison to other methods.} \it We compare to methods that report bounding-box part detection results, using their settings and measures. \vspace{-4mm} } 
\label{tab:comparisonOthers}
\end{table}

\subsection{Comparison to other part detection methods}
\label{sec:comparisons}

We compare here our full (bbox-reg) model (tab.~\ref{tab:resultsAlexNet}) to several prior works on detecting parts up to a bounding-box~\cite{chen14cvpr,zhang14eccv,gkioxari15iccv}. 
We use AlexNet, which is equivalent to the networks used in~\cite{zhang14eccv,zhang16cvpr,gkioxari15iccv}.
Tab.~\ref{tab:comparisonOthers} summarizes all results.

\vspT
\paragraph{Chen et al.~\cite{chen14cvpr}.} 
We compare to~\cite{chen14cvpr} following their protocol (sec.~4.3.3 of~\cite{chen14cvpr}).
They evaluate on 3 parts (head, body, and legs) of the 6 animal classes of PASCAL-Part, using Percentage of Correctly estimated Parts as measure (PCP).
They also need an extra measure called Percentage of Objects with Part estimated (POP), as they compute PCP only over object instances for which their system outputs a detection for that part class.
Additionally, they use ground-truth object bounding-boxes at test time.
More precisely, for each ground-truth box, they retain the best overlapping object detection, and evaluate part detection only within it.
As table~\ref{tab:comparisonOthers} shows, we outperform~\cite{chen14cvpr} on PCP, and our POP is substantially better, demonstrating the higher recall reached by our method.
We note how~\cite{chen14cvpr} only report results in this easier setting, whereas we report results in a fully automatic setting without using any ground-truth at test time (sec.~\ref{sec:resultsVal}).

\vspF
\paragraph{Fine-grained~\cite{zhang14eccv,lin15cvpr,zhang16cvpr}.} 
These fine-grained recognition works report part detection results on the CUB200-2011~\cite{WahCUB_200_2011} bird dataset
for the head and body.
They all evaluate using PCP and including object ground-truth bounding-boxes at test time.
Our model outperforms~\cite{zhang14eccv,lin15cvpr} by a large margin and is comparable to~\cite{zhang16cvpr}.
Only~\cite{zhang14eccv} report results without using object ground-truth at test time.
In this setting, our method performs almost as well as with object ground-truth at test time, achieving a very remarkable improvement (+25.8 PCP) compared to~\cite{zhang14eccv}.
Furthermore, we note that CUB200-2011 is an easier dataset than PASCAL-Part, with typically just one, large, fully visible bird instance per image.

\vspF
\paragraph{Gkioxari et al.~\cite{gkioxari15iccv}.}
This action and attribute recognition work reports detection results on three person parts (head, torso, legs) on PASCAL VOC 2009 images (tab.~1 in~\cite{gkioxari15iccv}).
As these do not have part ground-truth bounding-boxes, they construct them by grouping the keypoint annotations of~\cite{bourdev10eccv} (sec.~3.2.2 of~\cite{gkioxari15iccv}).
For an exact comparison, we train and test our full model using their keypoint-derived bounding-boxes and use their evaluation measure (AP at various IoU thresholds).
We also report (in parenthesis) results using the standard part ground-truth bounding-boxes of PASCAL-Part during both training and testing (as PASCAL VOC 2009 is a subset of PASCAL-Part).
We outperform~\cite{gkioxari15iccv} using their bounding-boxes, and obtain even better results using the standard bounding-boxes of PASCAL-Part.
Moreover, we note how their part detectors have been trained with more expensive annotations (on average 4 keypoints per part, instead of one bounding-box).

\subsection{Connection to part segmentation}
\vspO

Above we have considered directly detecting bounding-boxes on semantic parts.
Here we explore using a part segmentation technique as an intermediate step to obtain bounding-box detections, motivated by the apparently good performance of segmentation techniques~\cite{wang15iccv,hariharan15cvpr,liang16eccv,xia16eccv,chen17pami}.
We adapt a state-of-the-art part segmentation approach~\cite{chen17pami} by fitting bounding-boxes to its output segmentations.

We train DeepLab V2~\cite{chen17pami} with ResNet-101~\cite{he16cvpr} on all 105 parts of PASCAL-Part~\cite{chen14cvpr} \texttt{train} set, using the original pixel-wise annotations and following the training protocol of~\cite{chen17pami}.
At test time, we run the trained model on the \texttt{val} set of PASCAL-Part and place a bounding-box around each connected component of the output segmentation. 
This method achieves 11.6 mAP, averaged over all 105 part classes, which is much lower than what obtained by our full model, e.g. 40.1 mAP with VGG16 (despite ResNet-101 performing generally better than VGG16~\cite{he16cvpr}).

This exploratory experiment shows that it is not obvious how to go from segmentations to bounding-boxes. 
Powerful models such as DeepLab V2 have shown good results on a pixel-level part segmentation measure for a restricted set of classes (humans and quadrupeds~\cite{chen17pami}). However, these do not necessarily translate to good performance on an instance-level bounding-box measure (mAP), and when considering a more comprehensive set of classes.

\vspT
\subsection{Conclusions}
\vspT
We presented a semantic part detection model that detects parts in the context of their objects.
Our model includes several types of object information:
object class and appearance as indicators of what parts lie inside, and also part relative location conditioned on the object appearance.
Our model leads to a considerably better performance than detecting parts based only on their local appearance, improving by +5 mAP on the PASCAL-Part dataset.
Moreover, our model outperforms several other part detection methods~\cite{chen14cvpr,gkioxari15iccv,zhang14eccv,lin15cvpr} on PASCAL-Part and CUB200-2011.

\vspT
\paragraph{Acknowledgements.} This work was supported by the ERC Starting Grant VisCul.
A. Gonzalez-Garcia also acknowledges the Spanish project TIN2016-79717-R.

{\small
\bibliographystyle{unsrt}
\bibliography{../../bibtex/shortstrings,../../bibtex/vggroup,../../bibtex/calvin}
}

\end{document}